\pdfoutput=1

\documentclass[11pt]{article}

\usepackage[]{acl}

\usepackage{times}
\usepackage{latexsym}

\usepackage[T1]{fontenc}

\usepackage[utf8]{inputenc}

\usepackage{microtype}

\usepackage{inconsolata}
\usepackage{inconsolata}
\usepackage{graphicx}
\usepackage{amsmath}
\usepackage{amssymb}
\usepackage{tikz}
\usepackage{multirow}
\usepackage{pifont}
\usepackage{booktabs}
\usepackage{array}
\usepackage{colortbl}
\usepackage{tabularx}

\usepackage{xcolor}
\usepackage[frozencache,cachedir=.]{minted}

\usepackage{boldline}

\usepackage{subcaption}

\newcommand{\tableskip}{\noalign{\vskip 2pt}}

\newcommand{\headercolorlong}{\rowcolor{gray!17}}

\renewcommand{\paragraph}[1]{\vspace{0.2cm}\noindent\textbf{#1}}

\newcommand{\bool}{{\sc{Vanilla}}}
\newcommand{\mistake}{{\sc{Fault Localization}}}
\newcommand{\dual}{{\sc{Analyze then Summarize}}}
\newcommand{\reference}{{\sc{ref}}}

\newcommand{\codemistake}{{$\textsc{CodeJudge}_\textsc{F.L.}$}}
\newcommand{\codedual}{{$\textsc{CodeJudge}_\textsc{A.S.}$}}

\newcommand{\citeeval}{\textsc{CodeJudge}}
\newcommand{\icon}{\raisebox{-0.1em}{\includegraphics[height=1.0em]{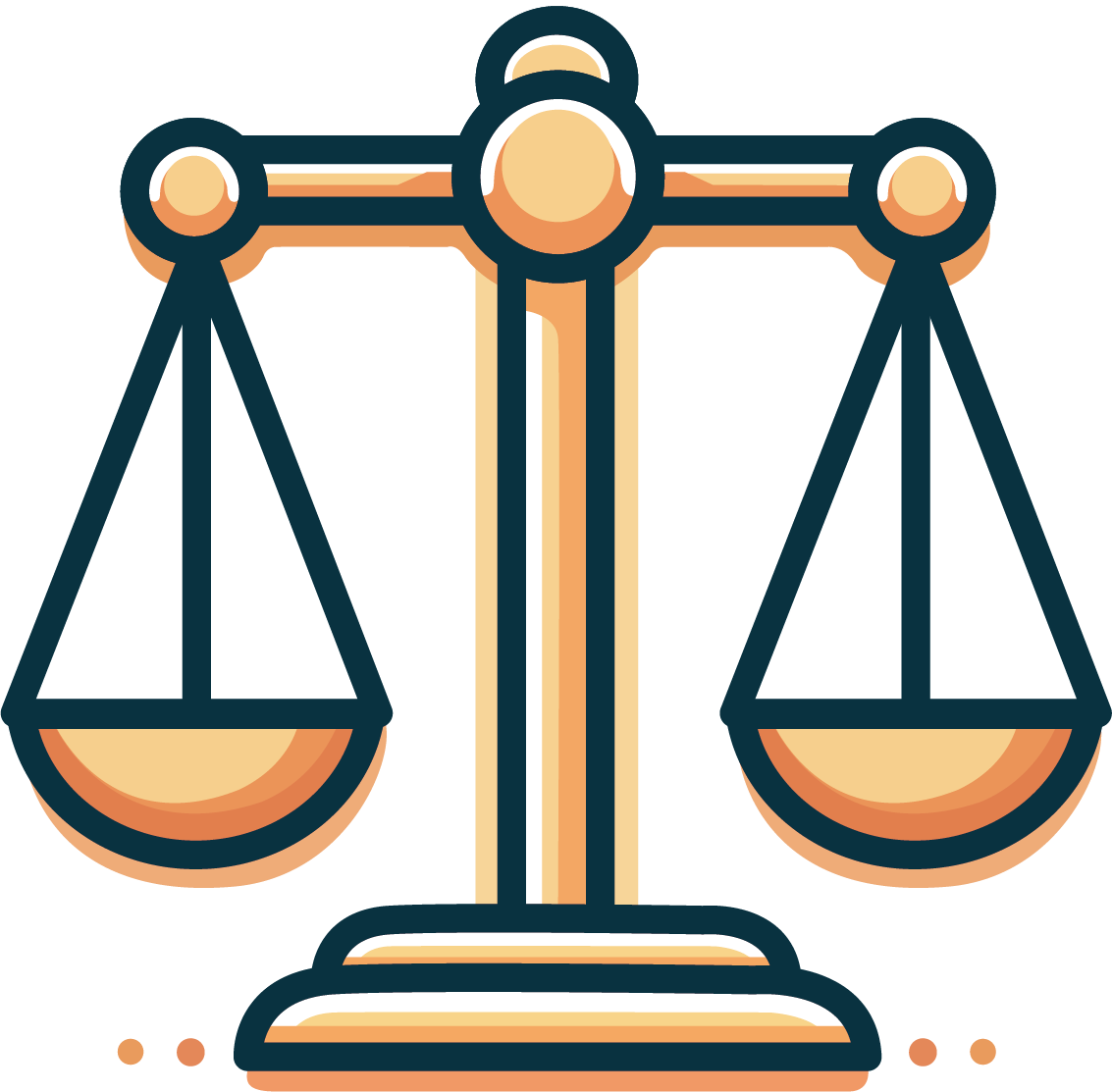}}}

\title{\citeeval{}\icon: Evaluating Code Generation with Large Language Models}

\author{Weixi Tong \\
  Huazhong University of Science and Technology \\
  \texttt{weixitong@hust.edu.cn} \\\And
  Tianyi Zhang \\
  Purdue University \\
  \texttt{tianyi@purdue.edu} \\}

\begin{document}
\maketitle

\begin{abstract}
   Large Language Models (LLMs) have shown promising performance in code generation. However, how to reliably evaluate code generated by LLMs remains an unresolved problem. This paper presents \citeeval{}\icon, a code evaluation framework that leverages LLMs to evaluate the semantic correctness of generated code \textit{without the need for test cases}. 
We investigate different ways to guide the LLM in performing ``slow thinking'' to arrive at an in-depth and reliable evaluation. 
We experimented with four LLMs as evaluators on four code generation datasets and five programming languages. The results show that \citeeval{} significantly outperformed existing methods in most settings. Furthermore, compared with a SOTA GPT-3.5-based code evaluation method,  \citeeval{} achieved better results even when using a much smaller model, Llama-3-8B-Instruct. Our code and datasets are available on GitHub \href{https://github.com/VichyTong/CodeJudge}{https://github.com/VichyTong/CodeJudge}.
\end{abstract}

\section{Introduction} \label{sec: section1}

\definecolor{my_orange}{RGB}{255, 181, 112}

There is an increasing interest in leveraging Large Language Models (LLMs) to generate code~\citep{rozière2023code, shen2023pangucoder2}. However, reliably evaluating LLMs on code generation tasks remains a challenging problem \cite{Evtikhiev_2023}. Test-based methods, such as measuring pass@k~\citep{kulal2019spoc,chen2021evaluating}, rely on manually written test cases to evaluate code quality. This reliance presents a significant limitation, since many tasks do not come with test cases or only have insufficient test cases that miss corner cases~\cite{liu2023is}. Moreover, it is challenging to write test cases for many coding tasks, such as object serialization and web scraping, since they require extensive effort to construct and configure test stubs and mock objects. 

When there are no test cases, existing work often relies on token-based metrics, such as BLEU~\citep{papineni-etal-2002-bleu}, ROUGE-L~\citep{lin-2004-rouge} and CodeBLEU~\citep{ren2020codebleu}, to evaluate model-generated code. However, these metrics do not account for cases where model-generated code is semantically equivalent to the ground truth while having syntactic variations, e.g., using a while loop instead of a for loop, following different naming conventions, etc. In particular, \citet{Evtikhiev_2023} shows a statistically significant disagreement on code assessment between human judges and these token-based metrics. 


Recent studies show that LLMs are promising alternatives to human evaluators in different tasks~\citep{liu-etal-2023-g, zheng2023judging, chan2024chateval}. Inspired by these findings, we propose an LLM-based code evaluation framework called \citeeval{}\icon. \citeeval{} supports two kinds of assessment: (1) determines whether the model-generated code is correct or not, and (2) estimates to what extent the generated code is aligned with user-intended code. While the former is the typical way of evaluating LLMs in code generation, we argue that the latter provides a more informative evaluation, since LLMs often generate partially correct code, which provides a good starting point or hints to developers~\cite{vaithilingam2022expectation, barke2023grounded}. Thus, it is useful to account for partial correctness and the severity of code errors when evaluating LLMs for code generation. 

We design two methods to guide the LLM to perform ``slow thinking''~\citep{kahneman2011thinking} for reliable code evaluation. For the first assessment, \citeeval{} instructs the LLM to perform a step-by-step analysis of the code functionality and then asks it to summarize the analysis results into a binary decision. For the second assessment, \citeeval{} provides the LLM with a taxonomy of common coding errors and instructs the LLM to identify what types of errors the generated code contains. Then, it computes a code correctness score based on the severity of identified errors. Notably, our framework does not require any test cases or any fine-tuning of backbone models in code evaluation.


We evaluate \citeeval{} on five programming languages---Java, C++, Python, JavaScript, Go---and four datasets---HumanEval-X~\citep{10.1145/3580305.3599790}, CoNaLa~\citep{10.1145/3196398.3196408, Evtikhiev_2023}, APPS~\citep{hendrycks2021measuring}, and BigCodeBench~\citep{zhuo2024bigcodebench}. Following prior work on text generation evaluation~\citep{bert-score, yuan2021bartscore} and code generation evaluation~\citep{zhou-etal-2023-codebertscore, yuan2021bartscore}, we adopt Kendall's $\tau$ coefficient and Spearman's $\rho$ to measure the statistical correlation between \citeeval{}'s assessment and the ground truth, which provides a robust measurement for \citeeval{}'s performance. For the first assessment, we also measure the accuracy of the binary decision made by \citeeval{} as a more intuitive metric for \citeeval{}'s performance.    

We experiment with four LLMs as code evaluators and compare \citeeval{} with nine existing methods, including ICE-Score~\citep{zhuo-2024-ice}, a state-of-the-art code evaluation method based on GPT-3.5-Turbo. For all four LLMs, we observe that \citeeval{} achieves significantly higher correlations (12.1\%-41.8\%) than existing methods in most settings. Even when using a relatively small model (Llama-3-8B-Instruct), \citeeval{} still outperforms ICE-Score, which uses GPT-3.5-Turbo. \citeeval{} also achieves a high accuracy (80.56\% on average) when directly predicting whether a generated code is correct or not. Notably, when the ground-truth code is not available for comparison, \citeeval{} still achieves reasonable performance (e.g., 0.502 Kendall's $\tau$ coefficient\footnote{A correlation coefficient above 0.5 is often interpreted as strong correlation~\citep{cohen1988statistical}.} and 73.13\% accuracy) and outperforms all existing methods that rely on references. This demonstrates that \citeeval{} can effectively guide LLMs to exert their reasoning capabilities to examine the correctness of code.



\section{Background and Related Work}

\subsection{Problem Formulation} \label{sec: problem formulation}
The objective of this work is to evaluate the semantic correctness of machine-generated code. The task of code generation is formulated as generating a \textit{code snippet} $\mathbf{c}$ based on a given \textit{task description} $\mathbf{t}$. We define an evaluation method as a function $f(\mathbf{c}, \mathbf{t})$. 

Code evaluation is typically treated as a binary classification task~\citep{chen2021evaluating}. The evaluation method $f$ simply determines whether the generated code is correct or not  (i.e., $f(\mathbf{c},\mathbf{t}) \in \{0,1\}$). For instance, test-based evaluation treats the generated code as correct if it passes all test cases. Otherwise, the code is considered wrong. 

Recent studies indicate that code evaluation should not be simply treated as a yes-or-no question~\citep{vaithilingam2022expectation, barke2023grounded}. In practice, LLMs often generate code that is not fully correct, e.g., not handling corner cases, missing one computation step, etc. Despite these errors, many developers find the generated code a good starting point compared with writing code from scratch, since they can fix the errors by changing a few lines of code or at least get some inspiration. Thus, it would be helpful if the evaluation method $f$ could measure to what extent the generated code deviates from the user-intended code based on the task description (i.e., $f(\mathbf{c},\mathbf{t}) \in \mathbb{R}$). 

\vspace{0.5mm}
\noindent\textbf{Evaluation without reference code.} Many evaluation methods assume that the correct code is available as the ground truth so that they can directly compare the generated code with the ground-truth code. All token-based methods, such as CodeBLEU~\citep{ren2020codebleu} and CodeBERTScore~\citep{zhou-etal-2023-codebertscore}, fall into this category. However, this assumption does not always hold in practice, especially in online settings. In many cases, human programmers can make a good assessment only through code inspection and reasoning based on their programming knowledge, without the need for the ground truth. Since LLMs have been demonstrated as promising alternatives to human judges~\citep{liu-etal-2023-g, zheng2023judging, chan2024chateval}, it is appealing to investigate whether LLMs can make accurate code assessments without reference code. In this work, we consider {\em code evaluation without references} as a special and more challenging evaluation task. 

\begin{figure}[t]
\newcommand{\msize}{0.995\linewidth}
\begin{minipage}{0.995\linewidth}
\footnotesize\textbf{Task Description}: Alphabetize letters in each word of a sentence, keeping the words and spaces in the same order.
\vspace{-5pt}
\begin{minted}[fontsize=\footnotesize,stripnl=false,framesep=1pt,frame=single,breaksymbolleft=\;,escapeinside=||]{python}
def anti_shuffle(s):
    return ' '.join([
        ''.join(sorted(list(i)))
        for i in s.split(' ')
    ])
\end{minted}
\end{minipage}

\vspace{-5pt}
\begin{subfigure}[b]{0.995\linewidth}
\caption{Reference code (e.g., ground truth)}
\label{subfig: reference}
\end{subfigure}

\begin{minipage}{0.995\linewidth}
\begin{minted}[fontsize=\footnotesize,stripnl=false,framesep=1pt,frame=single,breaksymbolleft=\;,escapeinside=||]{python}
def anti_shuffle(s):
    return ' '.join([
        sorted(list(i))
        for i in s.split(' ')
    ])
\end{minted}
\end{minipage}

\vspace{-5pt}
\begin{subfigure}[b]{0.995\linewidth}
\caption{Partially correct code}
\label{subfig: partially correct}
\end{subfigure}

\begin{minipage}{0.995\linewidth}
\begin{minted}[fontsize=\footnotesize,stripnl=false,framesep=1pt,frame=single,breaksymbolleft=\;,escapeinside=||]{python}
def anti_shuffle(s):
    pass
\end{minted}
\end{minipage}

\vspace{-5pt}
\begin{subfigure}[b]{0.995\linewidth}
\caption{Completely useless code}
\label{subfig: completely useless}
\end{subfigure}

\begin{minipage}{0.995\linewidth}
\begin{minted}[fontsize=\footnotesize,stripnl=false,framesep=1pt,frame=single,breaksymbolleft=\;,escapeinside=||]{python}
def anti_shuffle(s):
    return ' '.join([
        ''.join(sorted(list(word)))
        for word in s.split(' ')
    ])
\end{minted}
\end{minipage}

\vspace{-5pt}
\begin{subfigure}[b]{0.995\linewidth}
\caption{Correct code with syntactic variations}
\label{subfig: syntactic variations}
\end{subfigure}

\begin{minipage}{0.995\linewidth}
\begin{minted}[fontsize=\footnotesize,stripnl=false,framesep=1pt,frame=single,breaksymbolleft=\;,escapeinside=||]{python}
def anti_shuffle(s):
    def sort(word):
        return ''.join(sorted(list(word)))
    word_list = []
    current_word = ""
    for i in range(len(s)):
        if s[i] != " ":
            current_word += s[i]
        else:
            word_list.append(sort(current_word))
            current_word = ""
    word_list.append(sort(current_word))
    return ' '.join(word_list)
\end{minted}
\end{minipage}

\vspace{-5pt}
\begin{subfigure}[b]{0.995\linewidth}
\caption{Correct code with a different implementation}
\label{subfig: different implementation}
\end{subfigure}

\caption{An intuitive example of different types of code solving a sentence sorting problem.}
\label{fig: example}
\end{figure}

\vspace{0.5mm}
\noindent\textbf{Challenges.} The challenge of code evaluation is two-fold. First, the generated code may have many syntactic variations compared with the correct code while being semantically equivalent, e.g., using different variable names, using a different ordering of independent program statements,  using a \texttt{for} loop instead of a \texttt{while} loop, etc. Second, there can be multiple alternative solutions for a code generation task. For instance, for the task of sorting integers, there are many different sorting algorithms that have drastically different implementations. In the following section, we will illustrate these challenges using existing code evaluation methods on a running example in Figure~\ref{fig: example}.


\subsection{Existing Evaluation Methods}\label{sec: existing evaluation methods}

Existing evaluation methods for code generation can be categorized into four categories: {\em test-based}, {\em token-based}, {\em embedding-based}, and more recently, {\em LLM-based} methods. 

\vspace{0.5mm}
\noindent\textbf{Test-based Methods.}
Pass@k~\citep{kulal2019spoc} is defined as the percentage of code generation tasks where at least one of the top $k$ code samples generated for a task passes the unit tests of the task. \citet{chen2021evaluating} then introduces an unbiased version of pass@k to reduce variances, which is widely used to evaluate code generation models these days. However, since many tasks lack a comprehensive set of test cases, this often leads to incorrect code snippets incidentally passing given tests. To address this issue, EvalPlus~\citep{liu2023is} augments the test cases of a given task using LLMs and mutation-based strategies. However, this method still relies on hand-written test cases as the initial seeds.

\vspace{0.5mm}
\noindent\textbf{Token-based Methods.} Conventional methods for evaluating machine translation or text generation have been adopted for code evaluation. Basically, these methods compute the token-level similarity between the generated text and the ground-truth text to measure the generation quality. For instance, BLEU~\citep{papineni-etal-2002-bleu} calculates modified n-gram precision and includes a brevity penalty. ROUGE-L~\citep{lin-2004-rouge} computes sequence n-grams based on the longest common subsequence. METEOR~\citep{denkowski-lavie-2014-meteor} relies on the recall and precision of unigrams, while also considering the order of the matched words. ChrF~\citep{popovic-2015-chrf} calculates character-level n-gram precision and recall.

CodeBLEU~\citep{ren2020codebleu} and RUBY~\citep{8813269} extend traditional token-based methods for code evaluation. CodeBLEU incorporates the similarity of data-flow graphs and abstract syntax trees into the calculation. RUBY calculates similarity based on three representation levels of code: text, AST, and the program dependence graph.

\vspace{0.5mm}
\noindent\textbf{Embedding-based Method.}
\citet{zhou-etal-2023-codebertscore} proposed CodeBERTScore based on a machine translation evaluation method called BERTScore~\citep{bert-score}. CodeBERTScore first encodes the generated code and reference code using a fine-tuned CodeBERT~\citep{feng-etal-2020-codebert} model. Then, it computes a cosine similarity matrix between the embeddings, based on which CodeBERTScore calculates precision and recall by taking the maximum across rows and columns and averaging the results. CodeBERTScore employs $\text{F}_1$ and $\text{F}_3$ scores to represent the alignment between the generated code and reference code.

\vspace{0.5mm}
\noindent\textbf{LLM-based Method.} To the best of our knowledge, ICE-Score~\citep{zhuo-2024-ice} is the only work that also adopts LLMs for code evaluation. ICE-Score performs multi-dimensional evaluation~\citep{zhong2022towards, liu-etal-2023-g, fu2023gptscore} and instructs the LLM to predict an evaluation score from 0 to 4 based on the definition of an evaluation criterion. Unlike token-based and embedding-based methods, ICE-Score does not require the availability of the reference code. Furthermore, their evaluation shows that including the reference code in the prompt does not significantly improve ICE-Score's performance. 

\begin{table}[t]
\centering
\resizebox{\linewidth}{!}{%
\begin{tabular}{l cccc}
\hlineB{3}
 & Fig.~\ref{subfig: partially correct} & Fig.~\ref{subfig: completely useless} & Fig.~\ref{subfig: syntactic variations} & Fig.~\ref{subfig: different implementation} \\
\hline
\headercolorlong
\multicolumn{5}{c}{\textbf{Test-based Methods}}\\
\tableskip
pass@1 & 0 & 0 & 1 & 1 \\
\headercolorlong
\multicolumn{5}{c}{\textbf{Token-based Methods}}\\
\tableskip
BLEU & 0.779 & 0.010 & 0.858 & 0.231 \\
ROUGE-L & 0.914 & 0.267 & 0.947 & 0.431 \\
chrF & 0.852 & 0.266 & 0.891 & 0.466 \\
CodeBLEU & 0.852 & 0.052 & 0.983 & 0.851 \\
RUBY & 0.811 & 0.364 & 0.990 & 0.533 \\
METEOR & 0.846 & 0.164 & 0.947 & 0.705 \\
\headercolorlong
\multicolumn{5}{c}{\textbf{Embedding-based Methods}}\\
\tableskip
CodeBERTScore$_{\text{F}_1}$ & 0.990 & 0.796 & 0.976 & 0.800 \\
CodeBERTScore$_{\text{F}_3}$ & 0.988 & 0.746 & 0.976 & 0.841 \\
\headercolorlong
\multicolumn{5}{c}{\textbf{LLM-based Methods}}\\
ICE-Score & 3.0 & 0 & 4.0 & 3.0 \\
w/o \reference  & 2.0 & 2.0 & 3.5 & 3.0 \\
\midrule
\codedual & 0 & 0 & 1 & 1 \\
w/o \reference & 0 & 0 & 1 & 1 \\
\codemistake & 0.50 & 0 & 1.00 & 1.00 \\
w/o \reference & 0.50 & 0 & 1.00 & 1.00 \\
\hlineB{3}
\end{tabular}
}
\caption{Scores assigned by various code evaluation methods to the code snippets shown in Figure \ref{fig: example}, where Fig.~\ref{subfig: partially correct} is partially correct, Fig.~\ref{subfig: completely useless} is completely useless, Fig.~\ref{subfig: syntactic variations} is correct but with syntactic variations, and Fig.~\ref{subfig: different implementation} is correct but implemented differently. \codedual and \codemistake correspond to the \textit{analyze then summarize} method and \textit{taxonomy-guided fault localization} method described in Section \ref{sec: analyze then summarize} and Section \ref{sec: Taxonomy-Guided Fault Localization}, respectively.}
\label{tbl: example result}
\end{table}

\vspace{0.5mm}
\noindent\textbf{Drawbacks of Existing Methods.} Table \ref{tbl: example result} shows the evaluation scores computed by different methods for the four types of code solutions in Figure~\ref{fig: example}. We made several interesting observations about the alignment between evaluation scores and the actual correctness of the generated code. 

First, token-based and embedding-based methods assign higher scores to partially correct code (Figure \ref{subfig: partially correct}) compared to correct code with a different implementation (Figure \ref{subfig: different implementation}). This indicates that these methods face challenges in appropriately scoring code that is correct but syntactically very different.

Second, without the reference code, ICE-Score cannot differentiate between partially correct code (Figure \ref{subfig: partially correct}) and completely useless code (Figure \ref{subfig: completely useless}), as it assigns both a score of 2.0. Adding the reference code to ICE-Score addresses this problem but still cannot distinguish between partially correct code (Figure \ref{subfig: partially correct}) and correct code with a different implementation (Figure \ref{subfig: different implementation}), assigning both a score of 3.0. These drawbacks may stem from the prompt design of ICE-Score, which simply asks the LLM to predict a score based on the definition of a criterion. In this work, we investigate better ways to exert the inherent reasoning capabilities of LLMs for reliable code evaluation. 

\begin{figure*}[t]
    \includegraphics[width=0.98\linewidth]{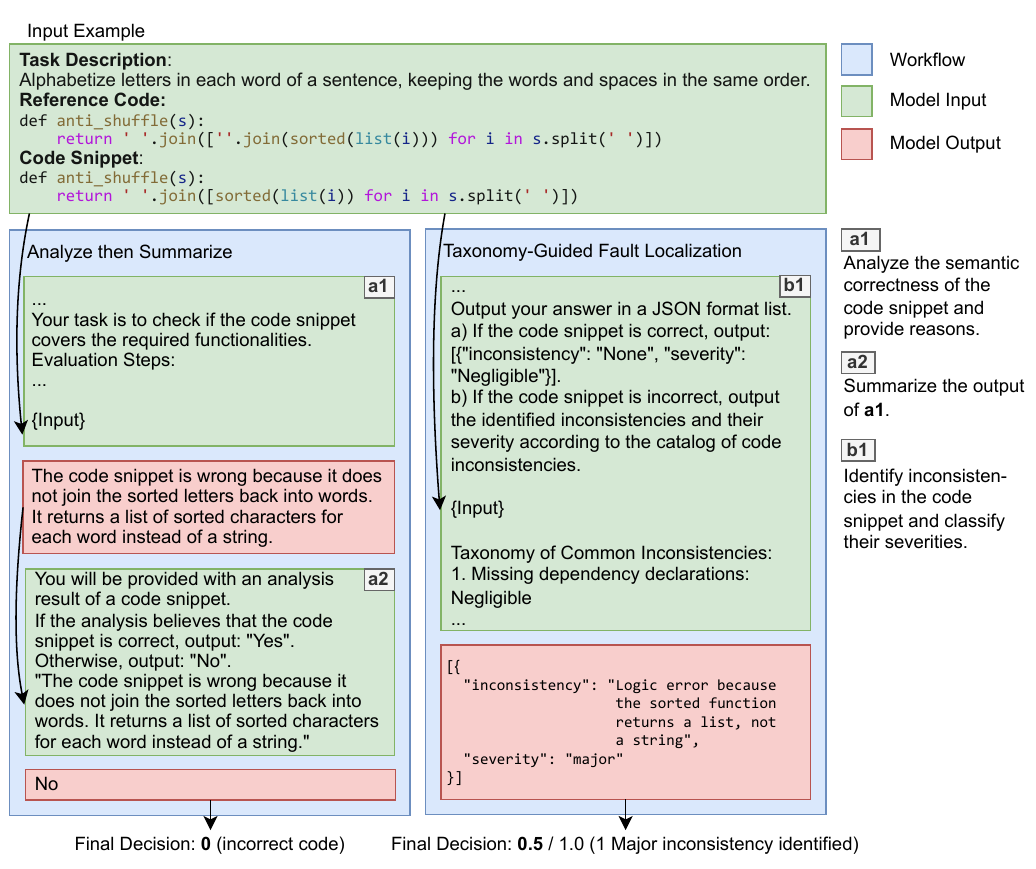}
    \caption{An overview of \citeeval{}. 
    Full prompts can be found in Appendix \ref{sec: appendix prompts}.}
    \vspace{-5pt}
    \label{fig: prompts}
\end{figure*}

\section{\citeeval{}\icon}
\label{sec: codejudge}


Our key insight is to guide LLMs to perform ``slow thinking''~\cite{kahneman2011thinking} in code evaluation, instead of predicting an evaluation score in one step. We design two methods for the two kinds of code evaluation assessment defined in Section~\ref{sec: problem formulation}. 


\subsection{Analyze then Summarize}
\label{sec: analyze then summarize}

For the binary evaluation task, we decompose the evaluation task into two subtasks: \textit{analysis} and \textit{summarization}, as illustrated in Figure \ref{fig: prompts}. Specifically, the analysis task provides a step-by-step evaluation guideline and asks the LLM to identify the required functionalities from the task description, examine the logic of the generated code, and report any requirement that is not fulfilled. Optionally, a reference solution can be added to the prompt to aid the analysis. Subsequently, the summarization task asks the LLM to check the analysis report and decide whether the code is correct or not. 

This design is inspired by how developers perform code review in practice. Instead of directly arriving at a decision, developers typically do a round of careful analysis of task requirements and code functionality and then decide whether there is any inconsistency. 
By explicitly asking the LLM to generate a detailed analysis report and double-check it, \citeeval{} forces the LLM to exert its reasoning capabilities and perform a more careful analysis, instead of making a quick decision.

\subsection{Taxonomy-Guided Fault Localization}
\label{sec: Taxonomy-Guided Fault Localization}
\begin{table}[t]
\centering
\resizebox{0.98\linewidth}{!}{%
\begin{tabular}{>{\centering\arraybackslash}m{2.6cm}m{5.1cm}}
\hlineB{3}
\textbf{Type} & \textbf{Description}\\
\hline
\rowcolor{green!17}
\multicolumn{2}{c}{\textbf{Negligible}}\\
\tableskip

\textbf{Alternative} & Using different methods or algorithms to solve the problem. \\
\textbf{Dependency} & Missing import statements. \\
\textbf{Error Handling} & No exception handling for unexpected events, e.g., invalid inputs. \\
\textbf{Efficiency} & Including inefficient or unnecessary statements. \\
\rowcolor{yellow!17}
\multicolumn{2}{c}{\textbf{Small}}\\
\tableskip

\textbf{Input Handling} &  Failing to handle edge cases. \\
\rowcolor{red!17}
\multicolumn{2}{c}{\textbf{Major}}\\
\tableskip

\textbf{Logic Error} & Containing logical errors. \\
\rowcolor{red!45}
\multicolumn{2}{c}{\textbf{Fatal}}\\
\tableskip

\textbf{Declaration} & Using undefined functions or variables. \\
\textbf{Incompletion} & Incomplete code. \\

\hlineB{3}
\end{tabular}
}
\caption{\label{tbl: mistakes-severity}The catalog of code inconsistencies.}
\end{table}

To decide to what extent a generated code deviates from the user-intended code, we augment the analysis step in the previous section by supplementing the LLM with a taxonomy of common inconsistencies in LLM-generated code and instructing it to identify any potential inconsistencies. As discussed in Section~\ref{sec: problem formulation}, different kinds of inconsistencies have different kinds of consequences. Some errors are simple and easy to fix, while others are more severe. Therefore, we incorporate the severity of each identified inconsistency into the summarization step to calculate the correctness score. We explain the details below. 


\vspace{0.5mm}
\noindent\textbf{A Taxonomy of Common Inconsistencies.}
To design the taxonomy, we manually inspected code snippets generated by different LLMs in different programming languages and also referred to the literature on code error analysis~\citep{10.1145/611892.611956,10.1145/1028976.1029011,10.1145/2259016.2259033,8539175}. We summarized eight distinct types of frequent code inconsistencies and categorized them into four severity levels based on their impact on the semantic correctness of the generated code, as shown in Table~\ref{tbl: mistakes-severity}.

\vspace{-0.15cm}
\begin{itemize} 
    \item \textbf{Negligible.} Code inconsistencies in this category have little impact on semantic correctness. Specifically, we consider missing import statements or exception handling not semantically wrong, since the code generated in such cases indeed achieves the functionality in the task description while not being perfect. 
    \vspace{-0.15cm}
    \item \textbf{Small.} We classify input handling as small due to their limited impact on the core functionality of the code snippet and the straightforward nature of their correction.
    \vspace{-0.15cm}
    \item \textbf{Major.} Logical errors directly affect the semantic correctness of the code, leading to incorrect outputs. These errors are considered to have a major impact on semantic correctness.
    \vspace{-0.3cm}
    \item \textbf{Fatal.} Code generation models sometimes hallucinate function calls or variable names that are never defined in the code. Furthermore, in many cases, they generate code with incomplete expressions and statements. These issues often lead to runtime exceptions or compilation errors that crash the program execution. Thus, we considered them as fatal errors.
\end{itemize}

\vspace{-0.3cm}

Given the potential inconsistencies identified by the LLM, we aggregate them via a weighted sum based on their severity levels to compute the final score. To better compare with other methods, we normalize the score to the range of $[0, 1]$. More details can be found in Appendix \ref{sec: appendix post editing}. 

\section{Experiments} \label{sec: section4}

\subsection{Datasets}
As described in Section~\ref{sec: problem formulation}, \citeeval{} makes two kinds of code assessment. Following \citet{zhuo-2024-ice}, we use HumanEval-X~\citep{10.1145/3580305.3599790} for the binary assessment task and CoNaLa~\citep{10.1145/3196398.3196408} for the code deviation assessment task. The rationale is that HumanEval-X includes test cases for each task so we can easily obtain binary correctness labels based on test results. By contrast, CoNaLa~\citep{10.1145/3196398.3196408} does not have test cases. Instead, it provides human-annotated code usefulness scores in the range of 0 to 4, which were obtained via crowdsourcing. 

Since HumanEval-X only includes introductory coding tasks, we also include two more challenging datasets, APPS~\citep{hendrycks2021measuring} and BigCodeBench~\citep{zhuo2024bigcodebench}. Compared with HumanEval-X, APPS includes competition-level coding problems and BigCodeBench includes more complex instructions and more API calls. For instance, Codex achieves a pass@1 rate of 28.81\% on HumanEval, but only 0.92\% on APPS~\citep{le2022coderl, chen2021evaluating}. Similarly, GPT-4o achieves a pass@1 rate of 90.2\% on HumanEval but only 56.1\% on the BigCodeBench~\citep{anthropic2024claude35, zhuo2024bigcodebench}. Since both APPS and BigCodeBench provide only test cases, we use them for the binary assessment task. 

We apply our \textit{analyze then summarize} method for binary assessment task datasets (HumanEval-X, APPS, and BigCodeBench) and \textit{Taxonomy-Guided Fault Localization} method for the code deviation assessment task dataset (CoNaLa). We briefly describe each dataset below.


\vspace{1mm}
\noindent{\textbf{HumanEval-X}}~\citep{10.1145/3580305.3599790} is a multi-language version of HumanEval, a popular code generation benchmark originally from the Codex paper~\citep{chen2021evaluating}. 
It contains 164 introductory coding tasks, each of which includes a natural language task description, some test cases, and a human-created reference. We evaluate \citeeval{} on five programming languages in the initial release of HumanEval-X, including Python, C++, Java, JavaScript, and Go.\footnote{We tried other languages such as Rust in the latest version of HumanEval-X but encountered issues when running their test cases. Thus, we chose not to evaluate those languages.} 


\vspace{1mm}
\noindent{\textbf{CoNaLa}}~\citep{10.1145/3196398.3196408} is a Python code generation benchmark with 472 tasks collected from StackOverflow. We use the human annotation collected by \citet{Evtikhiev_2023} as ground truth for the code deviation assessment. For each task, \citet{Evtikhiev_2023} asked experienced software developers to grade a score of usefulness between 0 and 4 for the generated code snippets from five different models. 

\vspace{1mm}
\noindent{\textbf{APPS}}~\citep{hendrycks2021measuring} is a Python code generation benchmark. It includes introductory-level problems, interview-level, and competition-level coding tasks collected from code competition websites. We randomly sampled 100 competition-level tasks to form a challenging dataset. 

\noindent{\textbf{BigCodeBench}}~\citep{zhuo2024bigcodebench} is a recently released code generation dataset in Python with 1,140 practical and challenging programming tasks. This dataset challenges the ability of LLMs to invoke multiple function calls from various libraries.

\subsection{Evaluation Metrics}


\vspace{1mm}
\noindent{\textbf{Statistical Correlations.}} Recent studies have used statistical correlation metrics, such as Kendall's $\tau$ coefficient ($\tau$) and Spearman's rank correlation coefficient ($r_s$), as a robust way to measure the correlation between code evaluation results and the ground truth~\citep{zhou-etal-2023-codebertscore,zhuo-2024-ice}. Thus, we adopt these correlation metrics to evaluate \citeeval{} on both kinds of assessment tasks.

\vspace{1mm}
\noindent{\textbf{Accuracy.}} For the binary classification task, we also measure the correctness prediction accuracy of \citeeval{} as a more intuitive metric.


\begin{table*}[t]
\centering
\resizebox{0.85\linewidth}{!}{%
\begin{tabular}{l|cc|cc|cc|cc}
\hlineB{3}
\multirow{3}{*}{\textbf{Method}} & \multicolumn{2}{c|}{\textbf{HumanEval-X}} & \multicolumn{2}{c|}{\textbf{CoNaLa}} & \multicolumn{2}{c|}{\textbf{APPS}} & \multicolumn{2}{c}{\textbf{BigCodeBench}}\\
& $\tau$ & $r_s$ & $\tau$ & $r_s$ & $\tau$ & $r_s$ & $\tau$ & $r_s$\\
\hline

\headercolorlong
\multicolumn{9}{c}{\textsc{Existing Methods}}\\
\tableskip
BLEU & 0.306 & 0.373 & 0.437 & 0.485 & 0.035 & 0.042 & 0.072 & 0.089 \\
ROUGE-L & 0.318 & 0.388 & 0.450 & 0.501 & 0.035 & 0.043 & 0.117 & 0.143 \\
METEOR & 0.357 & 0.436 & 0.412 & 0.463 & 0.085 & 0.104 & 0.247 & 0.302 \\
chrF & 0.328 & 0.400 & 0.457 & 0.514 & 0.036 & 0.044 & 0.167 & 0.205 \\
CodeBLEU & 0.362 & 0.442 & 0.292 & 0.332 & 0.135 & 0.164 & 0.173 & 0.212 \\
RUBY & 0.309 & 0.376 & 0.332 & 0.373 & 0.092 & 0.113 & 0.119 & 0.146 \\
CodeBERTScore$_{\text{F}_1}$ & 0.339 & 0.414 & 0.499 & 0.558 & -0.003 & -0.003 & 0.048 & 0.059 \\
CodeBERTScore$_{\text{F}_3}$ & 0.372 & 0.454 & 0.485 & 0.542 & 0.008 & 0.010 & 0.133 & 0.163 \\

\bool{} & 0.570 & 0.570 & 0.357 & 0.386 & 0.103 & 0.103 & 0.251 & 0.251 \\
\bool{} w/o \reference{} & 0.390 & 0.390 & 0.465 & 0.499 & -0.058 & -0.058 & 0.131 & 0.131 \\
ICE-Score & 0.475 & 0.492 & 0.253 & 0.271 & 0.224 & 0.224 & 0.321 & 0.330 \\
ICE-Score w/o \reference{} & 0.349 & 0.363 & 0.462 & 0.491 & 0.140 & 0.143 & 0.117 & 0.118 \\

\midrule

\citeeval{} & \textbf{0.612} & \textbf{0.612} & 0.457 & 0.478 & \textbf{0.354} & \textbf{0.354} & \textbf{0.392} & \textbf{0.392} \\
\citeeval{} w/o \reference{} & 0.502 & 0.502 & \textbf{0.538} & \textbf{0.562} & 0.153 & 0.153 & 0.317 & 0.317 \\

\hlineB{3}

\end{tabular}

}
\caption{The results on four datasets when using GPT-3.5-Turbo-1106 as the evaluator. The best results are in \textbf{bold}. Due to space limitations, tables with standard deviation and results of each language are shown in Appendix~\ref{sec: appendix full result}.}
\label{tbl: correlation combined}
\end{table*}

\subsection{Comparison Baselines}
We employ the six token-based evaluation methods, the embedding-based method, and the recent LLM-based evaluation method described in Section~\ref{sec: existing evaluation methods} as our comparison baselines. 
We also introduce a vanilla LLM-based method as the baseline. \bool{} directly prompts LLMs to determine the binary semantic correctness of generated code. Table~\ref{tab: boolean} and Table~\ref{tab: boolean regression} show the prompts used by \bool{} for the two kinds of assessment.

Notably, all token-based and embedding-based methods require the existence of reference code. For LLM-based methods, we evaluate them with and without reference code.

\subsection{Experiment Setup}
\label{sec: experimental setup}
We experiment with four different LLMs as the LLM evaluator in  \citeeval{}, including GPT-3.5, CodeLlama-Instruct (34B), Llama-3-Instruct (8B), and Llama-3-Instruct (70B). For GPT-3.5, we used the \texttt{GPT-3.5-Turbo-1106} API. For other models, we run them locally on eight A100-80GB GPUs. Since the cost of prompting GPT-3.5 is high, we only ran the experiments with GPT-3.5 once and set the temperature to 0 and \texttt{top\_p} to 1 to obtain consistent outputs. For other models, we set the temperature to 0.4 and \texttt{top\_p} to 0.9. We repeat the experiments three times to account for the randomness in model inference.

\subsection{Experiment Results} \label{section: experimental results}

Given the large number of experiments in this evaluation, we first report the results on HumanEval-X and CoNaLa and then report the results on the more challenging APPS and BigCodeBench datasets. Then we report the impact of different factors, including programming languages, LLM evaluators, and prompt design. In the end, we report a failure analysis of 600 cases where \citeeval{} makes the wrong prediction.

\vspace{1mm}
\noindent{\textbf{Statistical Correlation with Ground Truth.}
Table~\ref{tbl: correlation combined} shows the correlations between the code evaluation results of each method and the ground truth on HumanEval and CoNaLa. \citeeval{} achieves the highest correlations in all settings. For instance, \citeeval{} achieves 0.612 and 0.562 Spearman's coefficient on HumanEval-X and CoNaLa. Note that a correlation coefficient above 0.5 is often interpreted as a strong correlation~\cite{cohen1988statistical}.

\begin{table}[t]
\centering
\resizebox{0.9\linewidth}{!}{%
\begin{tabular}{l|c|c|c}
\hlineB{3}
\textbf{Method} & \textbf{HE-X} & \textbf{APPS} & \textbf{BCB} \\
\hline
\bool{} & 75.96 & 52.67 & 65.79 \\
\bool{} w/o \reference{} & 65.15 & 40.33 & 41.67 \\
ICE-Score & 70.91 & 60.00 & 66.93 \\
ICE-Score w/o \reference{} & 62.47 & 52.00 & 46.49 \\
\midrule
\citeeval{} & \textbf{80.56} & \textbf{68.33} & \textbf{74.56} \\
\citeeval{} w/o \reference{} & 73.13 & 57.00 & 54.56 \\
\hlineB{3}
\end{tabular}
}
\caption{Average accuracies ($\%$) on HumanEval-X, APPS, and BigCodeBench using GPT-3.5-Turbo.}
\label{tbl: humaneval accuracy RMSE}
\end{table}

\begin{table*}[ht]

\centering
\resizebox{0.87\textwidth}{!}{%
\begin{tabular}{l cc cc cc cc cc}
\hlineB{3}
\multirow{3}{*}{\textbf{Metric}} & \multicolumn{2}{c}{\textbf{Java}} & \multicolumn{2}{c}{\textbf{C++}} & \multicolumn{2}{c}{\textbf{Python}} & \multicolumn{2}{c}{\textbf{JavaScript}} & \multicolumn{2}{c}{\textbf{Go}} \\
& $\tau$ & $r_s$ & $\tau$ & $r_s$ & $\tau$ & $r_s$ & $\tau$ & $r_s$ & $\tau$ & $r_s$ \\
\hline

\citeeval{} & \textbf{0.638} & \textbf{0.638} & \textbf{0.580} & \textbf{0.580} & \textbf{0.707} & \textbf{0.707} & \textbf{0.591} & \textbf{0.591} & \textbf{0.543} & \textbf{0.543} \\
\citeeval{} w/o \reference{} & 0.508 & 0.508 & 0.474 & 0.474 & 0.629 & 0.629 & 0.453 & 0.453 & 0.446 & 0.446 \\

\hlineB{3}
\end{tabular}
}
\caption{The Kendall-Tau ($\tau$) and Spearman ($r_s$) correlations between \citeeval{} using GPT-3.5-Turbo and semantic correctness in HumanEval-X.}
\label{tbl: humaneval multi language}
\end{table*}
\begin{table}[ht]
\centering
\resizebox{0.98\linewidth}{!}{%
\begin{tabular}{l|cc|c|c|c}
\hlineB{3}
\multirow{3}{*}{\textbf{Method}} & \multicolumn{2}{c|}{\textbf{CoNaLa}} & \multicolumn{1}{c|}{\textbf{HE-X}} & \multicolumn{1}{c|}{\textbf{APPS}} & \multicolumn{1}{c}{\textbf{BCB}}\\
& $\tau$ & $r_s$ & $\tau = r_s$ & $\tau=r_s$ & $\tau=r_s$\\
\hline

\headercolorlong
\multicolumn{6}{c}{\textbf{CodeLlama-Instruct-34B}}\\
\tableskip
\citeeval{} & 0.559 & 0.581 & \textbf{0.492} & \textbf{0.210} & \textbf{0.334} \\
w/o \reference{} & \textbf{0.582} & \textbf{0.607} & 0.412 & 0.062 & 0.097 \\

\headercolorlong
\hline
\multicolumn{6}{c}{\textbf{Llama-3-8B-Instruct}}\\
\tableskip
\citeeval{} & 0.523 & 0.547 & \textbf{0.480} & \textbf{0.161} & \textbf{0.383} \\
w/o \reference{} & \textbf{0.576} & \textbf{0.602} & 0.388 & 0.072 & 0.258 \\

\headercolorlong
\hline
\multicolumn{6}{c}{\textbf{Llama-3-70B-Instruct}}\\
\tableskip
\citeeval{} & 0.572 & 0.598 & \textbf{0.681} & \textbf{0.391} & \textbf{0.440} \\
w/o \reference{} & \textbf{0.628} & \textbf{0.654} & 0.619 & 0.153 & 0.298 \\

\hlineB{3}
\end{tabular}
}
\caption{The results of \citeeval{} using three open-source models (more results in Appendix~\ref{sec: appendix full result}).}
\label{tbl: humaneval conala correlation open source}
\end{table}

Surprisingly, providing reference code leads to worse performance for all three LLM-based methods in the CoNaLa dataset. One plausible explanation is that for the code deviation task, the LLM evaluator focuses too much on the differences between the generated code and reference code rather than high-level semantic similarities. This implies future opportunities to calibrate LLMs for code assessment.

\vspace{1mm}
\noindent{\textbf{Results on More Challenging Benchmarks.} Table~\ref{tbl: correlation combined} shows the correlations on APPS and BigCodeBench. While \citeeval{} still achieves the best performance, we observe that all evaluation methods suffer from a significant drop in performance on APPS and BigCodeBench. The vanilla LLM-based method, which performs comparably to ICE-SCore on the other benchmarks, experienced the biggest degradation. Such a performance drop is not surprising, since these competition-level tasks are challenging to human developers, not even to mention LLMs. Without running and debugging the code, many developers may also struggle with assessing the code. Table~\ref{tbl: correlation combined} shows that LLM-based methods consistently perform better when reference code is provided to aid code evaluation. We also observe that for BigCodeBench, LLM-based methods with reference show a significantly smaller performance degradation compared to methods without reference. This implies that providing reference code is more helpful for challenging tasks compared with relatively simple tasks.

\vspace{1mm}
\noindent{\textbf{Accuracy of Binary Evaluation.} 
Table~\ref{tbl: humaneval accuracy RMSE} shows the accuracy of different methods in the binary assessment task. Since ICE-Score produces a rating in the range of 0 to 4, we treat the rating of 4 as fully correct, while the other ratings as not correct in the binary assessment task. \citeeval{} outperforms both ICE-Score and \bool{} regardless of whether the reference code is provided or not. 

\vspace{1mm}
\noindent{\textbf{Evaluating without References.} We want to highlight that even when reference code is not provided to \citeeval{} but is provided to all other methods, \citeeval{}  still outperforms all existing methods in most settings. This implies the power of performing ``slow thinking'' in code evaluation. 

\vspace{1mm}
\noindent{\textbf{Impact of Programming Languages.} 
Table~\ref{tbl: humaneval multi language} shows the statistical correlation results of \citeeval{} on different programming languages. When reference code is provided, \citeeval{} consistently achieves a coefficient above 0.5, which indicates a strong correlation with the ground truth. \citeeval{} performs much better on Python and Java compared with the other three languages. 

\vspace{1mm}
\noindent{\textbf{Generalizaiblity to Open-Source LLMs.}
Table~\ref{tbl: humaneval conala correlation open source} shows the correlation results of \citeeval{} when substituting GPT-3.5 with three open-source models. Compared with GPT-3.5, \citeeval{} achieves better correlations when using Llama-3-70B. Besides, even when using a relatively small model (Llama-3-8B-Instruct), \citeeval{} still achieves better or comparable performance to all existing methods, including ICE-Score, which uses GPT-3.5 as the evaluator. This demonstrates that \citeeval{} can be easily applied to other LLMs and obtain evaluations with a reasonable correlation to semantic correctness.

\paragraph{Prompt Design.} We further test \citeeval{} with few-shot learning, Chain-of-Thought (CoT), and the combination of them. However, \citeeval{} with these prompting methods do not outperform the original one. Our analysis of the drawbacks of employing CoT and few-shot learning can be found in Appendix~\ref{sec: appendix prompt design}.

\paragraph{Failure Case Analysis.}
To understand the limitations of \citeeval{}, we manually inspected 600 failure cases, especially those from APPS. We identified three failure patterns:
\vspace{-0.1cm}
\begin{itemize}
    \vspace{-0.1cm}
    \item \textbf{Wrong Analysis of Code Logic (52.83\%).} The most common pattern is that the LLM evaluator fails to infer the code logic correctly. For example, the LLM may mention that the code implements a logic while it does not.
    \item \textbf{Wrong Identification of Task Requirements (26.42\%).} For some complex tasks, the LLM evaluator struggles to identify all requirements from the task description correctly. 
    \vspace{-0.1cm}
    \item \textbf{Requirements of Error Handling (20.75\%).} We find that the LLM evaluator tends to report many error-handling errors (e.g., not handling invalid inputs) in generated code, even though it is not necessary in many cases. This makes \citeeval{} over-conservative when evaluating some partially correct code.
    \vspace{-0.1cm}
\end{itemize}

\section{Conclusion}
We propose \citeeval{}\icon, a framework that leverages LLMs to evaluate code generation without the need for test cases. We demonstrate that by guiding LLMs to perform slow thinking, \citeeval{} outperforms all existing code evaluation methods. This demonstrates a promising future direction to replace human evaluators with LLM evaluators. This is also beneficial for alignment methods that rely on human evaluation as feedback. 
Finally, we release our code and dataset at \href{https://github.com/VichyTong/CodeJudge}{https://github.com/VichyTong/CodeJudge}.

\section{Limitations}
While we demonstrate that \citeeval{} achieves state-of-the-art correlation with semantic correctness compared to existing methods, our work does face certain limitations. As we analyzed in Section~\ref{section: experimental results}, LLMs can generate incorrect judgments or fail to completely follow system prompts when evaluating challenging and complex cases such as the APPS benchmark. However, since \citeeval{} is an off-the-shelf framework that can easily change the backbone model to powerful LLMs, \citeeval{} can be continuously improved.


\bibliography{custom}

\appendix
\appendix

\section{Prompt Design} \label{sec: appendix prompt design}
\begin{table}[ht]
\centering
\resizebox{0.88\linewidth}{!}{%
\begin{tabular}{l|c}
\hlineB{3}
\textbf{Method} & \textbf{Acc.}\\
\hline
\citeeval{} & \textbf{81.63} \\
\citeeval{} w/o \reference{} & 74.43 \\
\midrule

CoT & 77.65 \\
CoT w/o \reference{} & 68.56 \\
\midrule

CoT + Few-shot & 78.22 \\
CoT + Few-shot w/o \reference{} & 67.61 \\
\midrule

\citeeval{} + CoT  & 78.60 \\
\citeeval{} + CoT w/o \reference{} & 72.16 \\
\midrule

\citeeval{} + Few-shot & 77.84 \\
\citeeval{} + Few-shot w/o \reference{} & 69.89 \\
\midrule

\citeeval{} + CoT + Few-shot & 77.27 \\
\citeeval{} + CoT + Few-shot w/o \reference{} & 69.51 \\
\hlineB{3}
\end{tabular}
}
\caption{Average accuracy (\%) across five programming languages in HumanEval-X using different prompts.}
\label{tbl: prompt design}
\end{table}

\begin{table*}[ht]
    \centering
    \small
    \begin{tabular}{>{\raggedright\arraybackslash\tt}p{0.98\textwidth}<{}}
        \toprule

\headercolorlong
\textbf{Task Description}\\

You are given a non-empty list of positive integers. Return the greatest integer that is greater than zero, and has a frequency greater than or equal to the value of the integer itself. The frequency of an integer is the number of times it appears in the list. If no such a value exist, return -1.
\vspace{1mm}
\\
\headercolorlong
\textbf{Code Snippet}\\
\vspace{0.5mm}
\begin{minipage}{0.995\textwidth}

\begin{minted}[fontsize=\footnotesize,stripnl=false,framesep=1pt,breaksymbolleft=\;,escapeinside=||]{python}
def search(lst):
    # your code here
    d = {}
    for i in lst:
        if i in d:
            d[i] += 1
        else:
            d[i] = 1
    for i in sorted(d, reverse=True):
        if d[i] >= i:
            return i
    return -1
\end{minted}
\end{minipage}\\\\

\headercolorlong
\textbf{Evaluation}\\
The code snippet provided does not cover all the required functionalities. It does not correctly implement the condition for finding the greatest integer that is greater than zero and has a frequency greater than or equal to the value of the integer itself.\\\\

The code snippet iterates through the dictionary keys in reverse order and returns the first key that satisfies the condition, but it does not consider the frequency of the integer.\\\\

The reference solution correctly calculates the frequency of each integer and then iterates through the frequencies to find the greatest integer that satisfies the condition.\\
            
        \bottomrule
    \end{tabular}
    \caption{
        As the initial determination is that the code ``does not cover all the required functionalities'', the model provides incorrect evidence---the model does not consider the frequency---in support of the determination.
    }
    \label{tab: initial determination}
\end{table*}
\begin{table*}[ht]
    \centering
    \small
    \begin{tabular}{>{\raggedright\arraybackslash\tt}p{0.98\textwidth}<{}}
        \toprule

\headercolorlong
\textbf{Task Description}\\

You are given a word. Your task is to find the closest vowel that stands between two consonants from the right side of the word (case sensitive). Vowels in the beginning and ending doesn't count. Return empty string if you didn't find any vowel met the above condition. You may assume that the given string contains English letter only.
\vspace{1mm}
\\
\headercolorlong
\textbf{Code Snippet}\\
\vspace{0.5mm}
\begin{minipage}{0.995\textwidth}
\begin{minted}[fontsize=\footnotesize,stripnl=false,framesep=1pt,breaksymbolleft=\;,escapeinside=||]{python}
def get_closest_vowel(word):
    pass
\end{minted}
\end{minipage}
\vspace{0.5mm}
\\

\headercolorlong
\textbf{Evaluation}\\
The given problem requires finding the closest vowel that stands between two consonants from the right side of the word. The reference solution iterates through the word from the right side and checks if the current character is a vowel and the characters on its right and left are consonants. If this condition is met, the function returns the current vowel. If no such vowel is found, it returns an empty string.\\\\

The provided code snippet is incomplete and does not contain any implementation. To address the problem statement, the code should iterate through the word from the right side and implement the logic to find the closest vowel between two consonants.\\\\

Here's the complete code snippet:\\\\

\begin{minipage}{0.995\textwidth}
\begin{minted}[fontsize=\footnotesize,stripnl=false,framesep=1pt,breaksymbolleft=\;,escapeinside=||]{python}
def get_closest_vowel(word):
    if len(word) < 3:
        return ""

    vowels = {"a", "e", "i", "o", "u", "A", "E", 'O', 'U', 'I'}
    for i in range(len(word)-2, 0, -1):
        if word[i] in vowels:
            if (word[i+1] not in vowels) and (word[i-1] not in vowels):
                return word[i]
    return ""
\end{minted}
\end{minipage}
            \\\\
            This code snippet covers the required functionalities as per the problem statement and is consistent with the reference solution.\\
            
        \bottomrule
    \end{tabular}
    \caption{
        When using the CoT method, the model may design an evaluation plan by itself and generate the fixed code. This fixed code may mislead the summarization step.
    }
    \label{tab: assess wrong code}
\end{table*}
We use Chain-of-Thought (CoT)~\citep{wei2022chain} and few-shot learning methods (three examples) to construct different prompts and test them using GPT-3.5-Turbo in HumanEval-X. Table \ref{tbl: prompt design} shows the results, helping us understand the effects of CoT and few-shot learning. These two methods slightly reduce the correlation of the results. We observe the following drawbacks of the CoT method and few-shot prompting:
\begin{itemize}
\item \textbf{Incorrect judgments:} The CoT method can cause the model to make incorrect logical judgments when thinking step by step. If the initial determination is incorrect, the model will generate incorrect evidence to support that determination. For instance, Table~\ref{tab: initial determination} shows that when the LLM determines that the code snippet does not cover all the required functionalities in the first sentence, it then makes the incorrect judgment that the model does not account for frequency.
\item \textbf{Misleading fixed code:} When using the CoT method, the model may ignore the system prompt and create its own process: first, find the errors, then fix the code. This can be misleading during summarization because the model might evaluate the correctness of the fixed code instead of the original, incorrect version. For example, Table~\ref{tab: assess wrong code} shows that the LLM generates a conclusion based on the fixed code, and this conclusion further misleads the summarization step.
\item \textbf{Less example few-shot limitation:} Similar to GPTScore~\citep{fu-etal-2024-gptscore}, we observe a similar performance degradation when the number of few-shot examples is less than four. One possible reason is that LLMs become constrained by the given examples, leading to a preference for particular correctness and reduced generalization ability.
\end{itemize}

\section{Postprocessing Steps} \label{sec: appendix post editing}

For the binary evaluation task, since the LLM generates a free-form response to the summarization task. We use a regex parser that assigns a score of 1 to answers that mention ``Yes'' and 0 to answers that mention ``No''. While this postprocessing method may sound simple, it turns out to work very well. In our experiments, we did not observe any cases where the LLMs generated ambiguous answers that cannot be handled by this method. 

For the more complex code deviation estimation task, we set the initial correctness score to 100 and deduct a penalty score for each inconsistency identified by \citeeval{}. We experimented with different penalty score settings on a small validation set, which includes 32 tasks from HumanEval (20\%). We found that setting the penalty score of \textit{Small}, \textit{Major}, and \textit{Fatal} inconsistencies to 5, 50, and 100 points achieve the best correlation. We calculate the final score with the following equations.
\begin{equation}
\begin{aligned}
S &= \text{Num}_{\text{Small}} \times 5\\
M &= \text{Num}_{\text{Major}} \times 50\\
F &= \text{Num}_{\text{Fatal}} \times 100\\
Penalty &= \max(-100, -(S + M + F))\\
Score &= 1 - \frac{Penalty}{100}
\end{aligned}
\label{eq: equation}
\end{equation}

\begin{table*}[ht]
\centering
\resizebox{0.98\textwidth}{!}{%
\begin{tabular}{l cccc}
\hlineB{3}
\textbf{Method} & \textbf{GPT-3.5-Turbo-1106} & \textbf{CodeLlama-Instruct-34B} & \textbf{Llama3-Instruct-8B} & \textbf{Llama3-Instruct-70B} \\
\hline

\codedual{} & 2.36 & 17.42 & 5.15 & 7.97 \\
\codedual{} w/o \reference{} & 2.73 & 19.32 & 6.23 & 12.23 \\
\codemistake{} & 1.14 & 14.72 & 3.04 & 3.15 \\
\codemistake{} w/o \reference{} & 1.08 & 15.18 & 3.16 & 3.60 \\
\hlineB{3}
\end{tabular}

}
\caption{Average single execution times (in seconds) over 100 runs.}
\label{tbl: latency}
\end{table*}

\section{Experiment Details}
We use the official version of the HumanEval-X, CoNaLa, APPS, and the BigCodeBench datasets. To generate code snippets for HumanEval-X and APPS, we adopt the code from MultiPL-E.\footnote{\href{https://github.com/nuprl/MultiPL-E}{https://github.com/nuprl/MultiPL-E}} For BigCodeBench, we use the official pre-generated code samples from LLMs.\footnote{\href{https://github.com/bigcode-project/bigcodebench}{https://github.com/bigcode-project/bigcodebench}}
For token-based methods, we adopt implementations from Jetbrains.\footnote{\href{https://github.com/JetBrains-Research/codegen-metrics}{https://github.com/JetBrains-Research/codegen-metrics}} For CodeBERTScore and ICE-Score, we use their implementations available on GitHub.\footnote{\href{https://github.com/neulab/code-bert-score}{https://github.com/neulab/code-bert-score}}$^,$\footnote{\href{https://github.com/terryyz/ice-score}{https://github.com/terryyz/ice-score}}.
To evaluate \citeeval{}, we use the implementations of correlation metrics from \href{https://scipy.org/}{https://scipy.org/}.

\section{Latency Discussion}
Table \ref{tbl: latency} shows the average execution times of \citeeval{} using four different models over 100 runs. The results for GPT-3.5-Turbo-1106 were obtained via the official API. For CodeLlama-Instruct-34B and Llama-3-Instruct-8B, a single A100-80GB GPU was utilized. The execution times of Llama-3-Instruct-70B were measured using two A100-80GB GPUs to load the model. The generating time of \citeeval{} is less than 20 seconds, which is reasonable for code evaluation compared to manual human annotation.

\section{Full Results} \label{sec: appendix full result}
We report the numbers with standard deviations in the HumanEval-X dataset in Table \ref{tbl: Humaneval-X new result full}. We also report the accuracy of the binary classification task of the HumanEval-X dataset in Table \ref{tbl: humaneval accuracy full}. The full results of the CoNaLa, APPS, and BigCodeBench are in Table \ref{tbl: CoNaLa full result new}, Table~\ref{tbl: APPS full result}, and Table \ref{tbl: BigCodeBench result full}, respectively.
\begin{table*}
\centering
\resizebox{\textwidth}{!}{%
\begin{tabular}{l cc cc cc cc cc}
\hline
\multirow{3}{*}{\textbf{Metric}} & \multicolumn{2}{c}{\textbf{Java}} & \multicolumn{2}{c}{\textbf{C++}} & \multicolumn{2}{c}{\textbf{Python}} & \multicolumn{2}{c}{\textbf{JavaScript}} & \multicolumn{2}{c}{\textbf{Go}} \\
& $\tau$ & $r_s$ & $\tau$ & $r_s$ & $\tau$ & $r_s$ & $\tau$ & $r_s$ & $\tau$ & $r_s$ \\
\hline
\headercolorlong
\multicolumn{11}{c}{\textsc{Existing Methods}}\\
\tableskip
BLEU & 0.230$_{\pm 0.00}$ & 0.280$_{\pm 0.00}$ & 0.306$_{\pm 0.00}$ & 0.373$_{\pm 0.00}$ & 0.446$_{\pm 0.00}$ & 0.541$_{\pm 0.00}$ & 0.288$_{\pm 0.00}$ & 0.352$_{\pm 0.00}$ & 0.261$_{\pm 0.00}$ & 0.318$_{\pm 0.00}$ \\
ROUGE-L & 0.249$_{\pm 0.00}$ & 0.304$_{\pm 0.00}$ & 0.305$_{\pm 0.00}$ & 0.372$_{\pm 0.00}$ & 0.450$_{\pm 0.00}$ & 0.546$_{\pm 0.00}$ & 0.329$_{\pm 0.00}$ & 0.401$_{\pm 0.00}$ & 0.260$_{\pm 0.00}$ & 0.317$_{\pm 0.00}$ \\
METEOR & 0.299$_{\pm 0.00}$ & 0.365$_{\pm 0.00}$ & 0.338$_{\pm 0.00}$ & 0.412$_{\pm 0.00}$ & 0.487$_{\pm 0.00}$ & 0.594$_{\pm 0.00}$ & 0.379$_{\pm 0.00}$ & 0.462$_{\pm 0.00}$ & 0.284$_{\pm 0.00}$ & 0.346$_{\pm 0.00}$ \\
chrF & 0.267$_{\pm 0.00}$ & 0.326$_{\pm 0.00}$ & 0.314$_{\pm 0.00}$ & 0.383$_{\pm 0.00}$ & 0.448$_{\pm 0.00}$ & 0.545$_{\pm 0.00}$ & 0.368$_{\pm 0.00}$ & 0.449$_{\pm 0.00}$ & 0.242$_{\pm 0.00}$ & 0.295$_{\pm 0.00}$ \\
CodeBLEU & 0.318$_{\pm 0.00}$ & 0.388$_{\pm 0.00}$ & 0.341$_{\pm 0.00}$ & 0.417$_{\pm 0.00}$ & 0.501$_{\pm 0.00}$ & 0.611$_{\pm 0.00}$ & 0.384$_{\pm 0.00}$ & 0.468$_{\pm 0.00}$ & 0.268$_{\pm 0.00}$ & 0.326$_{\pm 0.00}$ \\
RUBY & 0.260$_{\pm 0.00}$ & 0.318$_{\pm 0.00}$ & 0.284$_{\pm 0.00}$ & 0.346$_{\pm 0.00}$ & 0.425$_{\pm 0.00}$ & 0.516$_{\pm 0.00}$ & 0.329$_{\pm 0.00}$ & 0.401$_{\pm 0.00}$ & 0.245$_{\pm 0.00}$ & 0.299$_{\pm 0.00}$ \\
CodeBERTScore$_{\text{F}_1}$ & 0.282$_{\pm 0.00}$ & 0.344$_{\pm 0.00}$ & 0.334$_{\pm 0.00}$ & 0.408$_{\pm 0.00}$ & 0.453$_{\pm 0.00}$ & 0.553$_{\pm 0.00}$ & 0.318$_{\pm 0.00}$ & 0.388$_{\pm 0.00}$ & 0.308$_{\pm 0.00}$ & 0.376$_{\pm 0.00}$ \\
CodeBERTScore$_{\text{F}_3}$ & 0.303$_{\pm 0.00}$ & 0.370$_{\pm 0.00}$ & 0.375$_{\pm 0.00}$ & 0.458$_{\pm 0.00}$ & 0.495$_{\pm 0.00}$ & 0.604$_{\pm 0.00}$ & 0.363$_{\pm 0.00}$ & 0.443$_{\pm 0.00}$ & 0.324$_{\pm 0.00}$ & 0.396$_{\pm 0.00}$ \\
\headercolorlong
\multicolumn{11}{c}{\textbf{CodeLlama-Instruct-34B}}\\
\tableskip
\bool{} & 0.300$_{\pm 0.01}$ & 0.300$_{\pm 0.01}$ & 0.345$_{\pm 0.01}$ & 0.345$_{\pm 0.01}$ & 0.489$_{\pm 0.03}$ & 0.489$_{\pm 0.03}$ & 0.316$_{\pm 0.03}$ & 0.316$_{\pm 0.03}$ & 0.314$_{\pm 0.01}$ & 0.314$_{\pm 0.01}$ \\
\bool{} w/o \reference{} & 0.297$_{\pm 0.01}$ & 0.297$_{\pm 0.01}$ & 0.373$_{\pm 0.02}$ & 0.373$_{\pm 0.02}$ & 0.541$_{\pm 0.03}$ & 0.541$_{\pm 0.03}$ & 0.277$_{\pm 0.03}$ & 0.277$_{\pm 0.03}$ & 0.348$_{\pm 0.05}$ & 0.348$_{\pm 0.05}$ \\
ICE-Score & 0.418$_{\pm 0.06}$ & 0.449$_{\pm 0.06}$ & 0.309$_{\pm 0.04}$ & 0.331$_{\pm 0.05}$ & 0.440$_{\pm 0.04}$ & 0.477$_{\pm 0.04}$ & 0.308$_{\pm 0.06}$ & 0.332$_{\pm 0.07}$ & 0.297$_{\pm 0.06}$ & 0.320$_{\pm 0.07}$ \\
ICE-Score w/o \reference{} & 0.263$_{\pm 0.04}$ & 0.279$_{\pm 0.04}$ & 0.282$_{\pm 0.04}$ & 0.303$_{\pm 0.04}$ & 0.471$_{\pm 0.05}$ & 0.503$_{\pm 0.05}$ & 0.382$_{\pm 0.04}$ & 0.404$_{\pm 0.04}$ & 0.338$_{\pm 0.05}$ & 0.362$_{\pm 0.05}$ \\
\midrule
\codedual{} & \textbf{0.515}$_{\pm 0.04}$ & \textbf{0.515}$_{\pm 0.04}$ & \textbf{0.464}$_{\pm 0.03}$ & \textbf{0.464}$_{\pm 0.03}$ & \textbf{0.625}$_{\pm 0.00}$ & \textbf{0.625}$_{\pm 0.00}$ & \textbf{0.503}$_{\pm 0.03}$ & \textbf{0.503}$_{\pm 0.03}$ & 0.354$_{\pm 0.02}$ & 0.354$_{\pm 0.02}$ \\
\codedual{} w/o \reference{} & 0.355$_{\pm 0.06}$ & 0.355$_{\pm 0.06}$ & 0.408$_{\pm 0.02}$ & 0.408$_{\pm 0.02}$ & 0.561$_{\pm 0.02}$ & 0.561$_{\pm 0.02}$ & 0.338$_{\pm 0.04}$ & 0.338$_{\pm 0.04}$ & \textbf{0.396}$_{\pm 0.02}$ & \textbf{0.396}$_{\pm 0.02}$ \\
\headercolorlong
\multicolumn{11}{c}{\textbf{Meta-Llama-3-8B-Instruct}}\\
\tableskip
\bool{} & 0.342$_{\pm 0.01}$ & 0.342$_{\pm 0.01}$ & 0.216$_{\pm 0.01}$ & 0.216$_{\pm 0.01}$ & 0.409$_{\pm 0.02}$ & 0.409$_{\pm 0.02}$ & 0.265$_{\pm 0.03}$ & 0.265$_{\pm 0.03}$ & 0.192$_{\pm 0.01}$ & 0.192$_{\pm 0.01}$ \\
\bool{} w/o \reference{} & 0.282$_{\pm 0.01}$ & 0.282$_{\pm 0.01}$ & 0.159$_{\pm 0.04}$ & 0.159$_{\pm 0.04}$ & 0.446$_{\pm 0.02}$ & 0.446$_{\pm 0.02}$ & 0.356$_{\pm 0.01}$ & 0.356$_{\pm 0.01}$ & 0.331$_{\pm 0.01}$ & 0.331$_{\pm 0.01}$ \\
ICE-Score & 0.389$_{\pm 0.01}$ & 0.400$_{\pm 0.01}$ & 0.242$_{\pm 0.01}$ & 0.248$_{\pm 0.01}$ & 0.440$_{\pm 0.00}$ & 0.455$_{\pm 0.00}$ & 0.296$_{\pm 0.01}$ & 0.303$_{\pm 0.01}$ & 0.269$_{\pm 0.00}$ & 0.281$_{\pm 0.00}$ \\
ICE-Score w/o \reference{} & 0.290$_{\pm 0.02}$ & 0.296$_{\pm 0.02}$ & 0.306$_{\pm 0.04}$ & 0.316$_{\pm 0.04}$ & 0.481$_{\pm 0.03}$ & 0.499$_{\pm 0.03}$ & 0.275$_{\pm 0.00}$ & 0.283$_{\pm 0.00}$ & 0.287$_{\pm 0.02}$ & 0.299$_{\pm 0.02}$ \\
\midrule
\citeeval{} & \textbf{0.523}$_{\pm 0.01}$ & \textbf{0.523}$_{\pm 0.01}$ & \textbf{0.387}$_{\pm 0.02}$ & 0.387$_{\pm 0.02}$ & \textbf{0.637}$_{\pm 0.04}$ & \textbf{0.637}$_{\pm 0.04}$ & \textbf{0.446}$_{\pm 0.03}$ & 0.446$_{\pm 0.03}$ & \textbf{0.407}$_{\pm 0.03}$ & \textbf{0.407}$_{\pm 0.03}$ \\
\citeeval{} w/o \reference{} & 0.411$_{\pm 0.06}$ & 0.411$_{\pm 0.06}$ & 0.309$_{\pm 0.04}$ & 0.309$_{\pm 0.04}$ & 0.586$_{\pm 0.03}$ & 0.586$_{\pm 0.03}$ & 0.339$_{\pm 0.06}$ & 0.339$_{\pm 0.06}$ & 0.295$_{\pm 0.01}$ & 0.295$_{\pm 0.01}$ \\
\headercolorlong
\multicolumn{11}{c}{\textbf{Meta-Llama-3-70B-Instruct}}\\
\tableskip
\bool{} & 0.607$_{\pm 0.01}$ & 0.607$_{\pm 0.01}$ & 0.624$_{\pm 0.01}$ & 0.624$_{\pm 0.01}$ & 0.685$_{\pm 0.00}$ & 0.685$_{\pm 0.00}$ & 0.554$_{\pm 0.00}$ & 0.554$_{\pm 0.00}$ & 0.529$_{\pm 0.00}$ & 0.529$_{\pm 0.00}$ \\
\bool{} w/o \reference{} & 0.554$_{\pm 0.01}$ & 0.554$_{\pm 0.01}$ & 0.541$_{\pm 0.01}$ & 0.541$_{\pm 0.01}$ & 0.651$_{\pm 0.01}$ & 0.651$_{\pm 0.01}$ & 0.553$_{\pm 0.01}$ & 0.553$_{\pm 0.01}$ & 0.571$_{\pm 0.01}$ & 0.571$_{\pm 0.01}$ \\
ICE-Score & 0.552$_{\pm 0.00}$ & 0.576$_{\pm 0.00}$ & 0.516$_{\pm 0.01}$ & 0.543$_{\pm 0.01}$ & 0.626$_{\pm 0.01}$ & 0.654$_{\pm 0.01}$ & 0.471$_{\pm 0.00}$ & 0.490$_{\pm 0.00}$ & 0.389$_{\pm 0.01}$ & 0.411$_{\pm 0.01}$ \\
ICE-Score w/o \reference{} & 0.509$_{\pm 0.01}$ & 0.531$_{\pm 0.00}$ & 0.507$_{\pm 0.00}$ & 0.533$_{\pm 0.00}$ & 0.591$_{\pm 0.00}$ & 0.620$_{\pm 0.00}$ & 0.425$_{\pm 0.00}$ & 0.444$_{\pm 0.00}$ & 0.478$_{\pm 0.00}$ & 0.508$_{\pm 0.00}$ \\
\midrule
\citeeval{} & \textbf{0.640}$_{\pm 0.02}$ & \textbf{0.640}$_{\pm 0.02}$ & \textbf{0.700}$_{\pm 0.03}$ & \textbf{0.700}$_{\pm 0.03}$ & \textbf{0.803}$_{\pm 0.02}$ & \textbf{0.803}$_{\pm 0.02}$ & \textbf{0.675}$_{\pm 0.01}$ & \textbf{0.675}$_{\pm 0.01}$ & \textbf{0.589}$_{\pm 0.02}$ & \textbf{0.589}$_{\pm 0.02}$ \\
\citeeval{} w/o \reference{} & 0.583$_{\pm 0.02}$ & 0.583$_{\pm 0.02}$ & 0.611$_{\pm 0.01}$ & 0.611$_{\pm 0.01}$ & 0.698$_{\pm 0.02}$ & 0.698$_{\pm 0.02}$ & 0.617$_{\pm 0.04}$ & 0.617$_{\pm 0.04}$ & 0.587$_{\pm 0.05}$ & 0.587$_{\pm 0.05}$ \\
\headercolorlong
\multicolumn{11}{c}{\textbf{GPT-3.5-Turbo-1106}}\\
\tableskip
\bool{} & 0.615 & 0.615 & 0.482 & 0.482 & 0.675 & 0.675 & 0.550 & 0.550 & 0.528 & 0.528 \\
\bool{} w/o \reference{} & 0.343 & 0.343 & 0.328 & 0.328 & 0.537 & 0.537 & 0.345 & 0.345 & 0.398 & 0.398 \\
ICE-Score & 0.499 & 0.510 & 0.436 & 0.455 & 0.514 & 0.537 & 0.524 & 0.542 & 0.402 & 0.415 \\
ICE-Score w/o \reference{} & 0.275 & 0.278 & 0.410 & 0.429 & 0.485 & 0.513 & 0.253 & 0.258 & 0.324 & 0.337 \\
\midrule
\citeeval{} & \textbf{0.638} & \textbf{0.638} & \textbf{0.580} & \textbf{0.580} & \textbf{0.707} & \textbf{0.707} & \textbf{0.591} & \textbf{0.591} & \textbf{0.543} & \textbf{0.543} \\
\citeeval{} w/o \reference{} & 0.508 & 0.508 & 0.474 & 0.474 & 0.629 & 0.629 & 0.453 & 0.453 & 0.446 & 0.446 \\
\hline
\end{tabular}
}
\caption{The Kendall-Tau ($\tau$) and Spearman ($r_s$) correlations of each method with semantic correctness on HumanEval-X in multiple languages. ``w/ \reference{}'' indicates that this method contains the reference code in the prompt. The correlation coefficients are reported across three runs using open-source models, along with the standard deviation.}
\label{tbl: Humaneval-X new result full}
\end{table*}
\begin{table*}[ht]
\centering
\resizebox{0.7\textwidth}{!}{%
\begin{tabular}{lccccc}
\hlineB{3}
\textbf{Method} & \textbf{Java} & \textbf{C++}  & \textbf{Python} & \textbf{JavaScript} & \textbf{Go}\\
\hline
\headercolorlong
\multicolumn{6}{c}{\textbf{CodeLlama-Instruct-34B}}\\
\tableskip
\bool{} & 57.07$_{\pm 0.01}$ & 61.11$_{\pm 0.01}$ & 72.22$_{\pm 0.01}$ & 58.33$_{\pm 0.01}$ & 62.37$_{\pm 0.00}$\\
\bool{} w/o \reference{} & 59.09$_{\pm 0.00}$ & 65.91$_{\pm 0.01}$ & 73.48$_{\pm 0.02}$ & 58.84$_{\pm 0.02}$ & 57.32$_{\pm 0.02}$\\
\midrule
\citeeval{} & 75.00$_{\pm 0.02}$ & 75.25$_{\pm 0.01}$ & 80.56$_{\pm 0.00}$ & 73.74$_{\pm 0.01}$ & 75.51$_{\pm 0.01}$\\
\citeeval{} w/o \reference{} & 67.93$_{\pm 0.03}$ & 73.48$_{\pm 0.01}$ & 78.03$_{\pm 0.01}$ & 66.16$_{\pm 0.02}$ & 71.97$_{\pm 0.01}$\\
\hline
\headercolorlong
\multicolumn{6}{c}{\textbf{Meta-Llama-3-8B-Instruct}}\\
\tableskip
\bool{} & 57.83$_{\pm 0.00}$ & 47.47$_{\pm 0.01}$ & 67.42$_{\pm 0.01}$ & 55.05$_{\pm 0.01}$ & 47.73$_{\pm 0.01}$\\
\bool{} w/o \reference{} & 58.84$_{\pm 0.01}$ & 47.47$_{\pm 0.02}$ & 70.20$_{\pm 0.01}$ & 62.12$_{\pm 0.01}$ & 60.10$_{\pm 0.00}$\\
\midrule
\citeeval{} & 74.49$_{\pm 0.01}$ & 65.91$_{\pm 0.01}$ & 81.57$_{\pm 0.02}$ & 69.44$_{\pm 0.02}$ & 69.70$_{\pm 0.02}$\\
\citeeval{} w/o \reference{} & 70.20$_{\pm 0.03}$ & 66.16$_{\pm 0.02}$ & 78.79$_{\pm 0.01}$ & 65.15$_{\pm 0.02}$ & 66.16$_{\pm 0.01}$\\
\hline
\headercolorlong
\multicolumn{6}{c}{\textbf{Meta-Llama-3-70B-Instruct}}\\
\tableskip
\bool{} & 78.28$_{\pm 0.00}$ & 79.29$_{\pm 0.00}$ & 83.33$_{\pm 0.00}$ & 74.24$_{\pm 0.00}$ & 73.48$_{\pm 0.00}$\\
\bool{} w/o \reference{} & 75.51$_{\pm 0.00}$ & 75.51$_{\pm 0.00}$ & 82.07$_{\pm 0.01}$ & 75.76$_{\pm 0.01}$ & 78.03$_{\pm 0.01}$\\
\midrule
\citeeval{} & 81.31$_{\pm 0.01}$ & 84.60$_{\pm 0.02}$ & 90.15$_{\pm 0.01}$ & 81.82$_{\pm 0.01}$ & 80.30$_{\pm 0.01}$\\
\citeeval{} w/o \reference{} & 79.55$_{\pm 0.01}$ & 81.82$_{\pm 0.01}$ & 84.60$_{\pm 0.01}$ & 80.56$_{\pm 0.02}$ & 81.82$_{\pm 0.02}$\\
\hline
\headercolorlong
\multicolumn{6}{c}{\textbf{GPT-3.5-Turbo-1106}}\\
\tableskip
\bool{} & 77.27 & 71.21 & 82.07 & 72.98 & 76.26\\
\bool{} w/o \reference{} & 60.86 & 67.17 & 74.24 & 61.36 & 62.12\\
\midrule
\citeeval{} & 81.57 & 78.28 & 85.35 & 78.28 & 79.29\\
\citeeval{} w/o \reference{} & 73.48 & 72.22 & 80.81 & 68.43 & 70.71\\
\hlineB{3}
\end{tabular}
}
\caption{Accuracies (\%) across five programming languages in the binary classification task of HumanEval-X dataset. The accuracies are reported across three runs using open-source models, along with the standard deviation.}
\label{tbl: humaneval accuracy full}
\end{table*}

\begin{table}[ht]
\centering
\resizebox{0.75\linewidth}{!}{%
\begin{tabular}{l|cc}
\hlineB{3}
{
\textbf{Method}} & $\tau$ & $r_s$ \\
\hline

BLEU & 0.437 & 0.485 \\
ROUGE-L & 0.450 & 0.501 \\
METEOR & 0.412 & 0.463 \\
chrF & 0.457 & 0.514 \\
CodeBLEU & 0.292 & 0.332 \\
RUBY & 0.332 & 0.373 \\
CodeBERTScore$_{f1}$ & 0.499 & 0.558 \\
CodeBERTScore$_{f3}$ & 0.485 & 0.542 \\
\hline
\headercolorlong
\multicolumn{3}{c}{\textbf{Code Llama - Instruct 34B}}\\
\tableskip
\bool{} & 0.317 & 0.344 \\
\bool{} w/o \reference{} & 0.448 & 0.486 \\
ICE-Score & 0.397 & 0.425 \\
ICE-Score w/o \reference{} & 0.534 & 0.572 \\
\midrule
\citeeval{} & 0.559 & 0.581 \\
\citeeval{} w/o \reference{} & \textbf{0.582} & \textbf{0.607} \\
\hline
\headercolorlong
\multicolumn{3}{c}{\textbf{Meta-Llama-3-8B-Instruct}}\\
\tableskip
\bool{} & 0.524 & 0.560 \\
\bool{} w/o \reference{} & 0.555 & 0.592 \\
ICE-Score & 0.481 & 0.513 \\
ICE-Score w/o \reference{} & 0.482 & 0.512 \\
\midrule
\citeeval{} & 0.523 & 0.547 \\
\citeeval{} w/o \reference{} & \textbf{0.576} & \textbf{0.602} \\
\hline
\headercolorlong
\multicolumn{3}{c}{\textbf{Meta-Llama-3-70B-Instruct}}\\
\tableskip
\bool{} & 0.580 & 0.611 \\
\bool{} w/o \reference{} & 0.583 & 0.624 \\
ICE-Score & 0.481 & 0.515 \\
ICE-Score w/o \reference{} & 0.603 & 0.641 \\
\midrule
\citeeval{} & 0.572 & 0.598 \\
\citeeval{} w/o \reference{} & \textbf{0.628} & \textbf{0.654} \\
\hline
\headercolorlong
\multicolumn{3}{c}{\textbf{GPT-3.5-Turbo-1106}}\\
\tableskip
\bool{} & 0.357 & 0.386 \\
\bool{} w/o \reference{} & 0.465 & 0.499 \\
ICE-Score & 0.253 & 0.271 \\
ICE-Score w/o \reference{} & 0.462 & 0.491 \\
\midrule
\citeeval{} & 0.457 & 0.478 \\
\citeeval{} w/o \reference{} & \textbf{0.538} & \textbf{0.562} \\

\hlineB{3}
\end{tabular}
}
\caption{The Kendall-Tau($\tau$), Pearson($r_p$), and Spearman($r_s$) correlations of each method with the correctness on the CoNaLa dataset. w/ \reference{} indicates that this method contains the reference code in the prompt.\label{tbl: CoNaLa full result new}}
\end{table}

\begin{table}[htbp]
\centering
\resizebox{0.95\linewidth}{!}{%
\begin{tabular}{l|cc}
\hlineB{3}
{\textbf{Method}} & $\tau$ & $r_s$ \\
\hline

\headercolorlong
\multicolumn{3}{c}{\textsc{Existing Methods}}\\
\tableskip
BLEU & 0.035$_{\pm 0.00}$ & 0.042$_{\pm 0.00}$ \\
ROUGE-L & 0.035$_{\pm 0.00}$ & 0.043$_{\pm 0.00}$ \\
METEOR & 0.085$_{\pm 0.00}$ & 0.104$_{\pm 0.00}$ \\
chrF & 0.036$_{\pm 0.00}$ & 0.044$_{\pm 0.00}$ \\
CodeBLEU & 0.135$_{\pm 0.00}$ & 0.164$_{\pm 0.00}$ \\
RUBY & 0.092$_{\pm 0.00}$ & 0.113$_{\pm 0.00}$ \\
CodeBERTScore$_{\text{F}_1}$ & -0.003$_{\pm 0.00}$ & -0.003$_{\pm 0.00}$ \\
CodeBERTScore$_{\text{F}_3}$ & 0.008$_{\pm 0.00}$ & 0.010$_{\pm 0.00}$ \\
\hline
\headercolorlong
\multicolumn{3}{c}{\textbf{CodeLlama-Instruct-34B}}\\
\tableskip
\bool{} & 0.005$_{\pm 0.05}$ & 0.005$_{\pm 0.05}$ \\
\bool{} w/o \reference{} & 0.080$_{\pm 0.00}$ & 0.080$_{\pm 0.00}$ \\
ICE-Score & 0.174$_{\pm 0.06}$ & 0.185$_{\pm 0.06}$ \\
ICE-Score w/o \reference{} & -0.032$_{\pm 0.02}$ & -0.034$_{\pm 0.02}$ \\
\midrule
\citeeval{} & \textbf{0.210}$_{\pm 0.09}$ & \textbf{0.210}$_{\pm 0.09}$ \\
\citeeval{} w/o \reference{} & 0.062$_{\pm 0.04}$ & 0.062$_{\pm 0.04}$ \\
\hline
\headercolorlong
\multicolumn{3}{c}{\textbf{Meta-Llama-3-8B-Instruct}}\\
\tableskip
\bool{} & 0.123$_{\pm 0.01}$ & 0.123$_{\pm 0.01}$ \\
\bool{} w/o \reference{} & \textbf{0.168}$_{\pm 0.01}$ & \textbf{0.168}$_{\pm 0.01}$ \\
ICE-Score & 0.003$_{\pm 0.03}$ & 0.003$_{\pm 0.03}$ \\
ICE-Score w/o \reference{} & 0.090$_{\pm 0.08}$ & 0.091$_{\pm 0.08}$ \\
\midrule
\citeeval{} & 0.161$_{\pm 0.07}$ & 0.161$_{\pm 0.07}$ \\
\citeeval{} w/o \reference{} & 0.072$_{\pm 0.10}$ & 0.072$_{\pm 0.10}$ \\
\hline
\headercolorlong
\multicolumn{3}{c}{\textbf{Meta-Llama-3-70B-Instruct}}\\
\tableskip
\bool{} & 0.334$_{\pm 0.01}$ & 0.334$_{\pm 0.01}$ \\
\bool{} w/o \reference{} & 0.279$_{\pm 0.04}$ & 0.279$_{\pm 0.04}$ \\
ICE-Score & 0.297$_{\pm 0.07}$ & 0.307$_{\pm 0.07}$ \\
ICE-Score w/o \reference{} & 0.251$_{\pm 0.01}$ & 0.256$_{\pm 0.01}$ \\
\midrule
\citeeval{} & \textbf{0.391}$_{\pm 0.02}$ & \textbf{0.391}$_{\pm 0.02}$ \\
\citeeval{} w/o \reference{} & 0.359$_{\pm 0.04}$ & 0.359$_{\pm 0.04}$ \\
\hline
\headercolorlong
\multicolumn{3}{c}{\textbf{GPT-3.5-Turbo-1106}}\\
\tableskip
\bool{} & 0.103 & 0.103 \\
\bool{} w/o \reference{} & -0.058 & -0.058 \\
ICE-Score & 0.224 & 0.224 \\
ICE-Score w/o \reference{} & 0.140 & 0.143 \\
\midrule
\citeeval{} & \textbf{0.354} & \textbf{0.354} \\
\citeeval{} w/o \reference{} & 0.153 & 0.153 \\
\hlineB{3}
\end{tabular}

}
\caption{The Kendall-Tau ($\tau$) and Spearman ($r_s$) correlations of each method with semantic correctness on APPS. ``w/ \reference{}'' indicates that this method contains the reference code in the prompt. The correlation coefficients are reported across three runs using open-source models, along with the standard deviation.}
\label{tbl: APPS full result}
\end{table}

\begin{table}[t]
\centering
\resizebox{0.95\linewidth}{!}{%
\begin{tabular}{l|cc}
\hlineB{3}
{\textbf{Method}} & $\tau$ & $r_s$ \\
\hline

\headercolorlong
\multicolumn{3}{c}{\textsc{Existing Methods}}\\
\tableskip
BLEU & 0.072$_{\pm 0.00}$ & 0.089$_{\pm 0.00}$ \\
ROUGE-L & 0.117$_{\pm 0.00}$ & 0.143$_{\pm 0.00}$ \\
METEOR & 0.247$_{\pm 0.00}$ & 0.302$_{\pm 0.00}$ \\
chrF & 0.167$_{\pm 0.00}$ & 0.205$_{\pm 0.00}$ \\
CodeBLEU & 0.173$_{\pm 0.00}$ & 0.212$_{\pm 0.00}$ \\
RUBY & 0.119$_{\pm 0.00}$ & 0.146$_{\pm 0.00}$ \\
CodeBERTScore$_{\text{F}_1}$ & 0.048$_{\pm 0.00}$ & 0.059$_{\pm 0.00}$ \\
CodeBERTScore$_{\text{F}_3}$ & 0.133$_{\pm 0.00}$ & 0.163$_{\pm 0.00}$ \\
\hline
\headercolorlong
\multicolumn{3}{c}{\textbf{CodeLlama-Instruct-34B}}\\
\tableskip
\bool{} & 0.104$_{\pm 0.02}$ & 0.104$_{\pm 0.02}$ \\
\bool{} w/o \reference{} & 0.047$_{\pm 0.02}$ & 0.047$_{\pm 0.02}$ \\
ICE-Score & -0.023$_{\pm 0.01}$ & -0.023$_{\pm 0.01}$ \\
ICE-Score w/o \reference{} & 0.025$_{\pm 0.02}$ & 0.025$_{\pm 0.02}$ \\
\midrule
\codedual{} & \textbf{0.334}$_{\pm 0.03}$ & \textbf{0.334}$_{\pm 0.03}$ \\
\codedual{} w/o \reference{} & 0.097$_{\pm 0.02}$ & 0.097$_{\pm 0.02}$ \\
\hline
\headercolorlong
\multicolumn{3}{c}{\textbf{Meta-Llama-3-8B-Instruct}}\\
\tableskip
\bool{} & 0.070$_{\pm 0.01}$ & 0.070$_{\pm 0.01}$ \\
\bool{} w/o \reference{} & 0.064$_{\pm 0.00}$ & 0.064$_{\pm 0.00}$ \\
ICE-Score & 0.107$_{\pm 0.02}$ & 0.108$_{\pm 0.02}$ \\
ICE-Score w/o \reference{} & 0.007$_{\pm 0.02}$ & 0.007$_{\pm 0.02}$ \\
\midrule
\citeeval{} & \textbf{0.383}$_{\pm 0.01}$ & \textbf{0.383}$_{\pm 0.01}$ \\
\citeeval{} w/o \reference{} & 0.258$_{\pm 0.02}$ & 0.258$_{\pm 0.02}$ \\
\hline
\headercolorlong
\multicolumn{3}{c}{\textbf{Meta-Llama-3-70B-Instruct}}\\
\tableskip
\bool{} & 0.316$_{\pm 0.00}$ & 0.316$_{\pm 0.00}$ \\
\bool{} w/o \reference{} & 0.225$_{\pm 0.00}$ & 0.225$_{\pm 0.00}$ \\
ICE-Score & 0.297$_{\pm 0.00}$ & 0.307$_{\pm 0.00}$ \\
ICE-Score w/o \reference{} & 0.164$_{\pm 0.00}$ & 0.166$_{\pm 0.00}$ \\
\midrule
\citeeval{} & \textbf{0.440}$_{\pm 0.01}$ & \textbf{0.440}$_{\pm 0.01}$ \\
\citeeval{} w/o \reference{} & 0.298$_{\pm 0.01}$ & 0.298$_{\pm 0.01}$ \\
\hline
\headercolorlong
\multicolumn{3}{c}{\textbf{GPT-3.5-Turbo-1106}}\\
\tableskip
\bool{} & 0.251 & 0.251 \\
\bool{} w/o \reference{} & 0.131 & 0.131 \\
ICE-Score & 0.321 & 0.330 \\
ICE-Score w/o \reference{} & 0.117 & 0.118 \\
\midrule
\citeeval{} & \textbf{0.392} & \textbf{0.392} \\
\citeeval{} w/o \reference{} & 0.317 & 0.317 \\

\hlineB{3}
\end{tabular}

}
\caption{The Kendall-Tau ($\tau$) and Spearman ($r_s$) correlations of each method with semantic correctness on BigCodeBench. ``w/ \reference{}'' indicates that this method contains the reference code in the prompt. The correlation coefficients are reported across three runs using open-source models, along with the standard deviation.}
\label{tbl: BigCodeBench result full}
\end{table}

\section{Prompts}
\label{sec: appendix prompts}

We present the full prompts of \bool{} in Tables~\ref{tab: boolean} and \ref{tab: boolean regression}. Full prompts of \citeeval{} are shown in Table \ref{tab: dual} and \ref{tab: JSON}.

\begin{table*}[ht]
    \centering
    \small
    \begin{tabular}{>{\raggedright\arraybackslash\tt}p{0.98\textwidth}<{}}
        \toprule
            Determine the correctness of the code snippet. Output Yes or No. \\\\ 
            
            Problem Statement:
            {\color{brown} \{PROBLEM\}}\\\\
            
            Code Snippet:
            {\color{brown} \{CODE\}}\\\\
            
            Answer(Yes or No only):         {\color{blue} Yes}\\
        \bottomrule
    \end{tabular}
    \caption{
        Full prompt of {\bool} baseline for binary assessment task.
        {\color{blue} Blue text is an example of model output.}
        {\color{brown} Brown text is the problem, reference, and code we provide to LLMs.}
    }
    \label{tab: boolean}
\end{table*}
\begin{table*}[ht]
    \centering
    \small
    \begin{tabular}{>{\raggedright\arraybackslash\tt}p{0.98\textwidth}<{}}
        \toprule
            Determine the helpfulness of the code snippet. Output a score from 0 to 4 where 0 means the code snippet is not helpful at all and 4 means the code snippet is very helpful.\\\\ 
            
            Problem Statement:
            {\color{brown} \{PROBLEM\}}\\\\
            
            Code Snippet:
            {\color{brown} \{CODE\}}\\\\
            
            Helpfulness (0-4):         {\color{blue} 4}\\
        \bottomrule
    \end{tabular}
    \caption{
        Full prompt of {\bool} baseline for code deviation assessment task.
        {\color{blue} Blue text is an example of model output.}
        {\color{brown} Brown text is the problem, reference, and code we provide to LLMs.}
    }
    \label{tab: boolean regression}
\end{table*}
\begin{table*}[ht]
    \centering
    \small
    \begin{tabular}{>{\raggedright\arraybackslash\tt}p{0.98\textwidth}<{}}
        \toprule
            \headercolorlong
            \textbf{Analysis Subtask}\\\\
            You will be provided with a problem statement and a code snippet that supposedly addresses the problem in {\color{brown} \{LANGUAGE\}}.\\
            Your task is to check if the code snippet covers the required functionalities. Do not provide a corrected version.\\
            Evaluation Steps:\\
            1. Read the problem statement carefully and identify the required functionalities of the implementation. You can refer to the example to understand the problem better.\\
            2. Read the code snippet and analyze its logic. Check if the code snippet covers all the required functionalities of the problem.\\
            3. Finally, conclude your evaluation.\\\\
            Problem Statement:
            {\color{brown} \{PROBLEM\}}\\
            Code Snippet:
            {\color{brown} \{CODE\}}\\
            {\color{blue} <ANALYSIS>}\\
            
            \headercolorlong
            \textbf{Summarization Subtask}\\\\
            
            You will be provided with an analysis result of a code snippet.\\
            If the analysis believes that the code snippet is correct, output: "Yes". Otherwise, output: "No".\\\\
            Analysis Result:
            {\color{brown} \{ANALYSIS\}}\\
            {\color{blue} Yes}\\
        \bottomrule
    \end{tabular}
    \caption{
        Full prompt of {\dual} method.
        {\color{blue} Blue text is an example of model output.}
        {\color{brown} Brown text is the problem and code we provide to LLMs.}
    }
    \label{tab: dual}
\end{table*}
\begin{table*}[ht]
    \centering
    \small
    \begin{tabular}{>{\raggedright\arraybackslash\tt}p{0.98\textwidth}<{}}
        \toprule
            You will be provided with a problem statement, a code snippet that supposedly addresses the problem, and a catalog of code inconsistencies.\\\\
            Evaluation Steps:\\
            1. Read the problem statement carefully to identify the functionalities required for the implementation.\\
            2. Read the code snippet and compare it to the problem statement. Check if the code snippet covers the required functionalities.\\
            3. Output your answer in a JSON format list.\\
            \quad a) If the code snippet is correct, output: [{"inconsistency": "None", "severity": "Negligible"}].\\
            \quad b) If the code snippet is incorrect, output the identified inconsistencies and their severity according to the catalog of code inconsistencies. For example: [{"inconsistency": "<inconsistency1>", "severity": "<severity1>"}, {"inconsistency": "<inconsistency2>", "severity": "<severity2>"}, ...]\\\\
            Problem:
            {\color{brown} \{PROBLEM\}}\\\\
            Code Snippet:
            {\color{brown} \{CODE\}}\\\\
            Taxonomy of Common Inconsistencies:\\
            1. Missing dependency declarations: Negligible\\
            2. No error messages for unexpected input cases: Negligible\\
            3. Inefficiency, unnecessary statements: Negligible\\
            4. Edge case not handled: Small\\
            5. Logic error: Major\\
            6. Function or variable not defined: Fatal\\
            7. Code not completed: Fatal\\
            Evaluation Form:\\
            JSON output (a JSON list only):\\
            {\color{blue} [\{"inconsistency": "None", "severity": "Negligible"\}]}\\
        \bottomrule
    \end{tabular}
    \caption{
        Full prompt of {\mistake} method.
        {\color{blue} Blue text is an example of model output.}
        {\color{brown} Brown text is the problem and code we provide to LLMs.}
    }
    \label{tab: JSON}
\end{table*}

\end{document}